\documentclass[sigconf]{acmart}

\usepackage{xcolor,colortbl}
\definecolor{c1}{HTML}{D1D17A}
\definecolor{c2}{HTML}{C0A545}
\definecolor{c3}{HTML}{C27E3A}
\definecolor{c4}{HTML}{C47557}
\definecolor{c5}{HTML}{B05F3C}

\definecolor{c6}{HTML}{99E0FF}
\definecolor{c7}{HTML}{BBBBFF}
\definecolor{c8}{HTML}{CD85FE}
\definecolor{c9}{HTML}{C79BF2}
\definecolor{c10}{HTML}{A6CAA9}
\definecolor{c11}{HTML}{8EB4E6}
\definecolor{c12}{HTML}{FFC48E}
\definecolor{c13}{HTML}{EEF093}
\definecolor{c14}{HTML}{C88E8E}

\definecolor{rgb1}{rgb}{0.5725490196078431, 0.7764705882352941, 1.0}

\definecolor{rgb2}{rgb}{0.592156862745098, 0.9411764705882353, 0.6666666666666666}

\definecolor{rgb3}{rgb}{1.0, 0.6235294117647059, 0.6039215686274509}

\definecolor{rgb4}{rgb}{0.8156862745098039, 0.7333333333333333, 1.0}

\definecolor{rgb5}{rgb}{1.0, 0.996078431372549, 0.6392156862745098}

\definecolor{rgb6}{rgb}{0.6901960784313725, 0.8784313725490196, 0.9019607843137255}
\usepackage{booktabs} % For formal tables
\usepackage{cleveref}
\usepackage{xspace}
\usepackage{soul}
\usepackage{balance}

\newcommand{\helix}{{\sc Helix}\xspace}

\newcommand{\name}{\helix}

\setcopyright{none}

\newenvironment{denselist}{
    \begin{list}{\tiny{$\bullet$}}%
    {\setlength{\itemsep}{0ex} \setlength{\topsep}{0ex}
    \setlength{\parsep}{0pt} \setlength{\itemindent}{0pt}
    \setlength{\leftmargin}{1.5em}
    \setlength{\partopsep}{0pt}}}%
    {\end{list}}

\newcommand{\topic}[1]{\vspace{-3.5pt}\smallskip \smallskip \noindent{\bf #1.}}
\newcommand{\hidden}[1]{}

\renewcommand\footnotetextcopyrightpermission[1]{}
\begin{document}
\title{How Developers Iterate on Machine Learning Workflows}
\subtitle{A Survey of the Applied Machine Learning Literature}

%%%%%%%%%%  Authors %%%%%%%%%%%%%%%%

\author{Doris Xin, Litian Ma, Shuchen Song, Aditya Parameswaran}
\affiliation{%
  \institution{University of Illinois,  Urbana-Champaign (UIUC)}
}
\email{{dorx0, litianm2, ssong18, adityagp}@illinois.edu}

%%%%%%%%%      End of Authors       %%%%%%%%%%%%%

\renewcommand{\shortauthors}{D. Xin et al.}

% force figs to be on different pages
\renewcommand{\floatpagefraction}{0.1}

\begin{abstract}
%!TEX root=main.tex
Machine learning workflow development 
is anecdotally regarded 
to be an iterative process of trial-and-error with humans-in-the-loop.
However, we are not aware of statistics-based evidence 
corroborating this popular belief.
A statistical characterization of iteration 
can serve as a  benchmark for machine learning workflow development in practice, 
and can aid the development of 
human-in-the-loop machine learning systems.
To this end, we conduct a small-scale survey of 
the applied machine learning literature 
from five distinct application domains. 
We use statistics collected from the papers 
to estimate the role of iteration within machine learning workflow development,
and report preliminary trends and insights from our investigation,
as a starting point towards this benchmark.
Based on our findings,
we finally describe desiderata for 
effective and versatile human-in-the-loop machine learning
systems that can cater to users in diverse domains.

\end{abstract}

\maketitle

\section{Introduction}
\label{sec:intro}
%!TEX root=main.tex

Development of machine learning (ML) applications is governed by an iterative process:
starting with an initial workflow,
developers iteratively modify their workflow, based on previous results,
to improve performance.
They may add or modify data sources, features,
hyperparameters, and training algorithms, among others.
These iterations of trial-and-error are necessary due to 
data variability, algorithmic complexity, and overall
unpredictability of ML.
A {\em detailed, statistical characterization of 
how developers iteratively modify ML workflows 
can serve as a benchmark for 
human-in-the-loop ML systems}. 
At present,
due to the lack of such studies,
we are forced to resort to anecdotal evidence
to identify usage patterns and motivate design decisions. 

To this end, {\em we conduct a statistical study
of iteration by surveying the applied ML literature
across five application domains}.
The statistics collected in this study provide
the first quantitative evidence of how developers
iterate on ML workflows, beyond anecdotal ones.
Moreover, the insights and trends discovered from our survey
provide concrete guidelines on desired 
human-in-the-loop ML system properties,
while the models and statistics provide a starting point
for the development of benchmarks for standardized
and automatic evaluation of human-in-the-loop ML systems.

Statistical studies of end-to-end ML workflow development 
pose several challenges.
First, it is difficult to gather 
data that captures the entire process, and not just
the final snapshot. 
One approach, for example, may involve examining code
repositories over time
to determine what has changed---one 
downside of this approach is that 
developers may not commit intermediate iterations, 
leading to less transparency for the overall process.
Moreover, this approach will require
understanding code, and mapping code fragments to
classes of iterative modifications,
both of which are extremely challenging to do. 
Second, we need to ensure that our study
captures a diverse set of application domains. 
Surveys~\cite{oreilly17, kaggle17, mitreview17,munson2012study} often end up focusing on 
industry-relevant application areas (e-commerce, recommendations),
and data-types (language, vision). 
Since our eventual goal is to develop a benchmark for 
general-purpose human-in-the-loop
ML systems, this limited view may 
hinder our ability to adequately support all application domains.
Third, once the data is collected, 
we need to devise methods to analyze the data 
and collect statistics related to iteration.
Finally, we need to turn the raw statistics into models that capture iteration
and relate trends and insights discovered from these models to ML system design.

Our study includes an analysis of 
105 applied machine learning papers
sampled from multiple conferences in 2016
and across five application domains, 
including social sciences,
 natural sciences, 
web application, 
computer vision, 
and natural language processing.
We collect statistics from each paper that capture iterative development
and use these statistics to infer
common practices in each application domain surveyed.
We describe the statistics collected, 
how they are used to estimate iteration counts, and
discuss the limitations of our approach in the next section.
To ensure the quality of our statistics, 
we take consensus over results collected by multiple surveyors,
and open-source the final aggregated 
data for further studies by interested readers,
as well as development of formal benchmarks.
We conduct data analysis on our survey results 
to highlight key insights unearthed by our survey
and propose system requirements suggested by our analysis.

\topic{Related Work}
To the best of our knowledge, our survey is the first effort in 
conducting a statistical study of machine learning model 
development from empirical evidence.
However, the pursuit of understanding iterative ML development is not singularly ours.
Several surveys have been conducted in recent years to profile industry and academic ML users~\cite{oreilly17, kaggle17, mitreview17,munson2012study}.
These surveys differ from ours
in that they were self-reported responses from a 
select set of industry and academic users.
Findings from self-reporting surveys are known to suffer from response bias~\cite{nederhof1985methods}.
Many articles discuss general trends and design patterns 
in ML workflows~\cite{domingos2012few, Jordan255, royalsociety17},
while a number of articles focus on 
providing guidance and taxonomies for novice users to perform iteration better~\cite{bestPractice, qiu2016survey, gartner17}.
Other works such as~\mbox{\cite{abdul2018trends}} and~\mbox{\cite{liu2014chi}}
study general trends and needs in data science 
using NLP techniques to study a large corpus en masse.
Vartak et al.~\cite{vartak2015supporting} describe
a system-building vision for iterative human-in-the-loop ML.
Kery et al.~\cite{kery2017variolite} specifically study
the versioning aspect of iterative development, 
whereas Koesten et al.~\mbox{\cite{koesten2017trials}} analyze 
in-depth surveys to understand the typical workflow for data scientists.

\smallskip
\noindent
The rest of paper is organized as follows: 
In Section~\ref{sec:data}, 
we describe the data, 
the statistics collected from the data, 
and the methods to study iteration using the statistics. 
In Section~\ref{sec:results}, 
we report interesting results and insights discovered from our survey 
and propose concrete system requirements 
to support human-in-the-loop ML based on the survey analysis.

\section{Data \& Methodology}
\label{sec:data}
%!TEX root=main.tex

In this section we describe the dataset and the methods 
used to collect the statistics that enable analyses of iteration in publications.

\subsection{Corpus}
\label{sec:corpus}

{We surveyed 105 papers published in 2016 on applied data science.
To ensure relevance, 
we selected four venues that 
specifically publish applied machine learning studies:
KDD Applied Data Science Track,
Association for Computational Linguistics (ACL),
Computer Vision and Pattern Recognition (CVPR),
and Nature Biotechnology (NB).
We randomly sample 20 papers from ACL, CVPR, and NB each,
and 45 papers from KDD.
These papers span
applications in social sciences (SocS), web applications (WWW), natural sciences (NS), natural language processing (NLP), and computer vision (CV). 
Paper topics were determined using the ACM Computing Classification System (CCS)~\footnote{https://www.acm.org/publications/class-2012}.
Keywords in each paper are matched with entries in the CCS tree, 
and each paper is assigned as its domain 
the most appropriate high level entry containing its keywords.
Figure~\ref{fig:domain} illustrates the domain composition of the conferences surveyed. 
While ACL, CVPR, and Nature specialize in a single domain,
KDD embraces many domains, with a focus on web applications and social science.

\begin{figure}
\vspace{-12pt}
\centering
\includegraphics[width=0.4\textwidth]{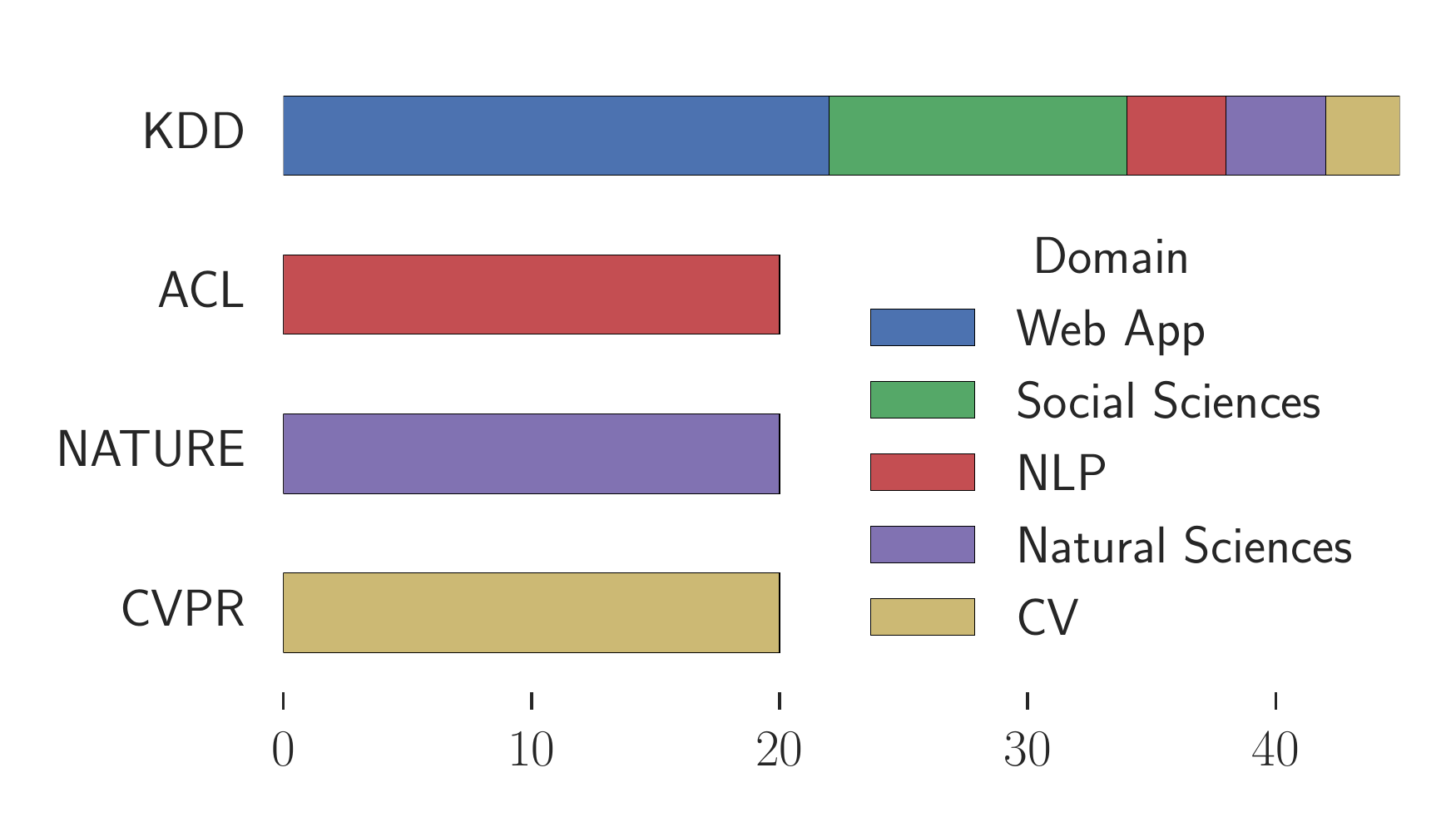}
\vspace{-15pt}
\caption{Paper count per domain by conference.}
\label{fig:domain}
\end{figure}

\topic{Limitations}
Our approach is limited in its ability to accurately model iterations 
due to several characteristics of the corpus:
\begin{denselist}
\item[1)] While the corpus spans multiple domains, 
the number of paper in each domain is small,
which can lead to spurious trends.
\item[2)] Papers provide an incomplete picture of the overall iterative process.
Machine learning papers are results-driven 
and focus more on modeling than data pre-processing by convention.
Due to space constraints, authors often omit a large number of iterative steps
and report only on the small subset that led to the final results.
\item[3)] Papers often present results side by side instead of the order they were obtained,
making it difficult to determine the exact transitions 
between the variants studied in the iterative process.
\end{denselist}
We attempt to overcome some of these limitations by 
\begin{denselist}
\item Having multiple surveyors and aggregating the results to reduce the change of spurious results, to be elaborated in Section~\ref{sec:stat};
\item Devising estimators that do no rely on information about the order of operations,
to be elaborated in Section~\ref{sec:infer}.
\end{denselist}

\subsection{Brief Overview of ML Workflows}
ML workflows commonly consist of three major components:

\topic{Data Pre-processing (DPR)}
This stage contains all the data manipulation operations,
such as data cleaning and feature extraction,
used to turn raw data into a format compatible with ML algorithms.

\topic{Learning/Inference (L/I)}
Once the data is transformed into a learnable representation,
such as feature vectors,
learning takes place, using the transformed data to derive an ML model via optimization.
Inference refers to the processing by which the learned model is used to make predictions on unseen data, and is often performed after learning.

\topic{Post Processing (PPR)}
Post processing is the all-encompassing term for operations following learning and inference.
Bruha et al. \cite{bruha2000postprocessing} classifies PPR operations in to four categories:
1) rule-based knowledge filtering, 
2) and knowledge integration,
3) interpretation and explanation,
4) evaluation.
While 1) and 2) involve transformations of the L/I output,
3) and 4) are about the analysis of the L/I output.
Mentions of 1) and 2) are sparse in our corpus 
and thus excluded from our study.

In the context of ML application development, 
an {\em iteration} involves creating a version of the workflow,
either from scratch or by copying/modifying a previous version,
and executing this version end to end to obtain some results.
Program termination marks the end of an iteration,
and any results that are not written to disk during execution
can only be obtained by modifying the workflow to explicitly save the results
and rerunning the workflow.

\subsection{Statistics Collection}
\label{sec:stat}
Our goal in this survey is to collect statistics on how users iterate on ML workflows.
However, iterations are often not explicitly reported in publications. 
To overcome this challenge, we design a set of statistics
that allow us to infer the iterative process 
leading to the results reported in each paper.
We introduce the statistics for each individual component of the ML workflow below.

\topic{DPR}
%\dorx{Maybe expand this out to the full list if there is room.}
As mentioned above, DPR encompasses all operations 
involved in transforming raw data into learnable representations,
such as feature engineering, data cleaning, and feature value normalization.
We record $\mathcal{D}$, the set of distinct DPR operation types found in each paper
and collect $n_{\mathcal{D}} = |\mathcal{D}|$.
Mentions of DPR operations are usually found in the data and methods sections in the paper.

\topic{L/I}
Workflow modifications concerning L/I fall into one of three categories:
1) hyperparameter tuning for a model (e.g., increasing learning rate, changing the architecture of a neural net) and
2) switching between model classes (e.g., from decision tree to SVM).
For each paper, we record $\mathcal{M}$, the set of all model classes and
$\mathcal{P}$, the set of distinct hyperparameters tuned across all model classes,
and collect $n_{\mathcal{M}} = |\mathcal{M}|$ and
$n_{\mathcal{P}} = |\mathcal{P}|$.
Evidence for these statistics is usually found in the algorithms section,
as well as result tables and figures.

\topic{PPR}
Of the four types of PPR operations enumerated above, 
evaluation and interpretation/explanation are the most commonly reported in papers,
often presented in tables or figures.
For each paper, we record $\mathcal{E}$, the set of evaluation metrics used,
and collect $n_{\mathcal{E}} = | \mathcal{E} |$.
In addition, we collect $n_{table}$ and $n_{figure}$, 
the number of tables and figures containing results and case studies, respectively.

We refer to $\mathcal{D}, \mathcal{M}, \mathcal{P}, \mathcal{E}$  collectively as \textit{entity sets} in the rest of the paper~\footnote{The complete entity sets and statistics can be found at https://github.com/gestalt-ml/AppliedMLSurvey/blob/master/data/combinedCounts.tsv}.

\vspace{2pt}
To ensure the quality of the statistics collected, 
we had three graduate students in data mining,
henceforth referred to as \textit{surveyors}, 
perform the survey independently on the same corpus.
We reference the results collected by each surveyor with a subscript,
e.g., $\mathcal{M}_1$ is the set of model classes recorded by surveyor 1.
To increase the likelihood of consensus, 
we first had the surveyors discuss and agree on 
a seed set for each entity set, 
e.g., $\mathcal{E} =$ \{Accuracy, RMSE, NDCG\}.
Surveyors were then asked to remove from and add to this set as they see fit for each paper.
Let $n'_x$ be the aggregated value of the statistic $n_x$.
We aggregate the three sets of results as follows:
\begin{denselist}
\item For an entity set $S$ (e.g., $\mathcal{M}$, the set of model classes),
let $S_a = S_1 \cup S_2 \cup S_3$.
We filter $S_a$ to obtain $S' \subseteq S_a$ 
such that $s \in S'$ is identified by at least two surveyors.
That is, a paper is considered to contain an operation 
only if it is identified to be in the paper by at least two surveyors independently.
We define $n'_S$ for the corresponding statistic as $|S'|$. 
\item For $n_{table}$ and $n_{figure}$, 
we define $n'_{table/figure}$ to be the average of the values obtained by the three surveyors.
\end{denselist}

\subsection{Estimating Iterations using Statistics}
\label{sec:infer}
The information collected above indicate 
versions of the workflow studied but not the iterative modifications themselves.
To infer the number of iterations using the statistics collected above,
we make the following assumptions:
\begin{denselist}
\item Each iteration involves a single change.
While it is possible for multiple changes to be tested in a single iteration,
it is unlikely the case since the interactions can obfuscate the contribution of individual changes.
\item Each element in an entity set is tested exactly once.
For the authors to report on a variant, 
there must have been at least one version of the workflow containing that variant.
Although it is likely for a variant to be revisited in multiple iterations in the actual research process,
papers, by convention, provide little information on this aspect.
Due to this lack of evidence, we take the conservative approach by taking the minimum value.
\end{denselist}

\noindent Let $t_{DPR}, t_{LI}, t_{PPR}$ be the number of iterations 
containing changes to the DPR, L/I, and PPR components of the workflow, respectively.
Using the two assumptions above, 
we estimate $t_{DPR}, t_{LI}$, and $t_{PPR}$ as follows:
\begin{denselist}
\item $\hat{t}_{DPR} = n'_{\mathcal{D}}$
\item $\hat{t}_{LI} = (n'_{\mathcal{M}} - 1) + (n'_{\mathcal{P}} - 1)$
\item $\hat{t}_{PPR} = \min \left(n'_{\mathcal{E}}, n'_{table} + n'_{figure} \right)$
\end{denselist}
For $\hat{t}_{DPR}$, we assume that 
the authors start with the raw data
and incrementally add more data pre-processing operations in each iteration.
We subtract one from $n'_{\mathcal{M}}$ and $n'_{\mathcal{P}}$
in  $\hat{t}_{LI}$ to account for the fact that 
the initial version of the workflow must contain 
a model, a set of hyperparameters, and an optimization algorithm.
The estimator $\hat{t}_{PPR}$ assumes that in a PPR iteration, 
the authors can either gather all information on a single metric or generate an entire figure/table.

\section{Results and Insights}
\label{sec:results}
%!TEX root=main.tex

In this section we share interesting trends about ML workflow development 
discovered from our survey.

\subsection{Iteration Count}
\label{sec:iterCount}

\begin{figure}[h]
\centering
\includegraphics[width=0.5\textwidth]{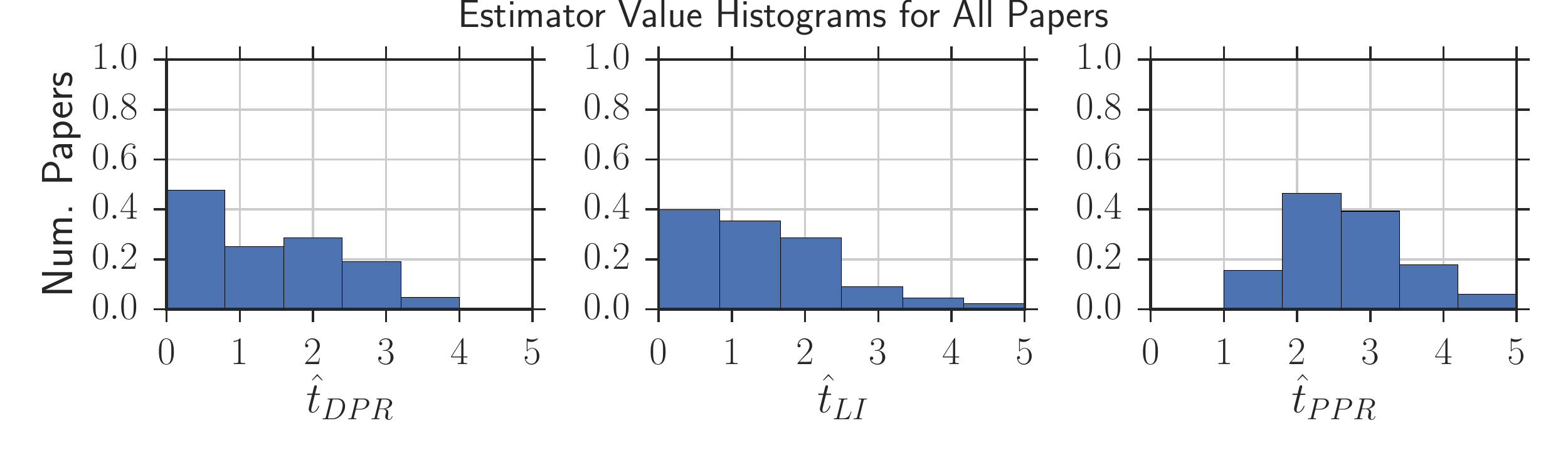}
\includegraphics[width=0.5\textwidth]{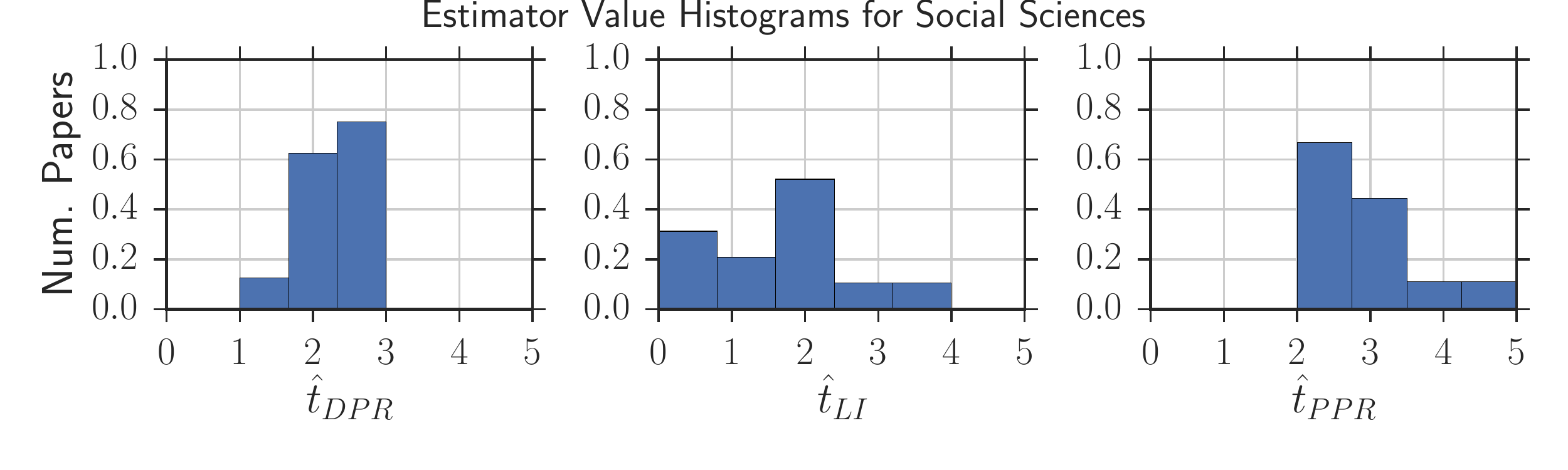}
\includegraphics[width=0.5\textwidth]{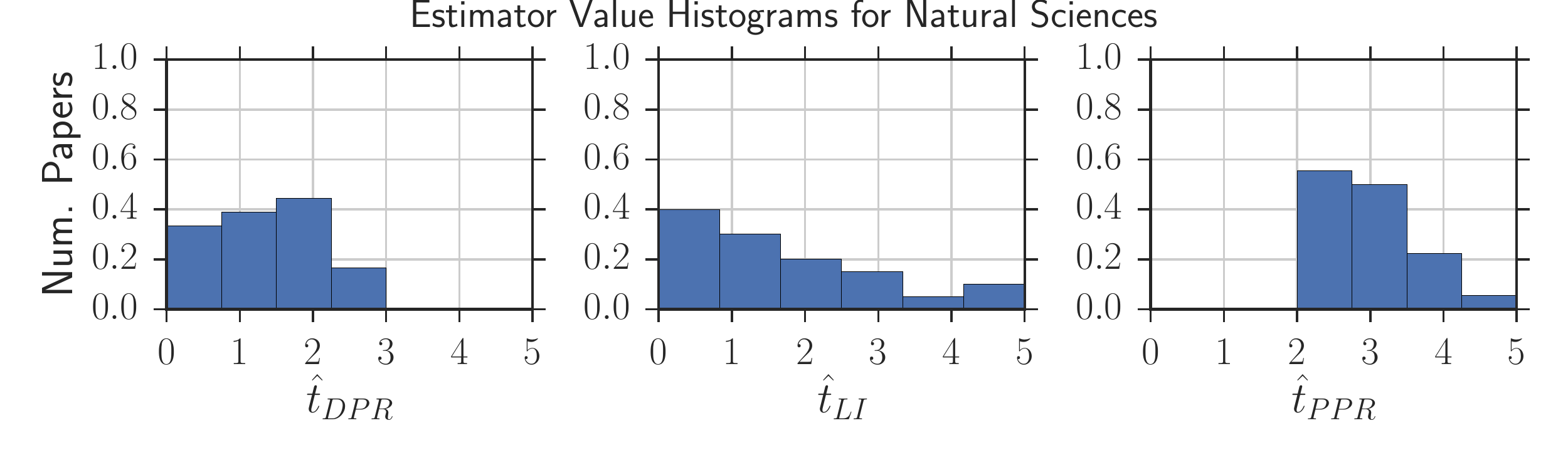}
\includegraphics[width=0.5\textwidth]{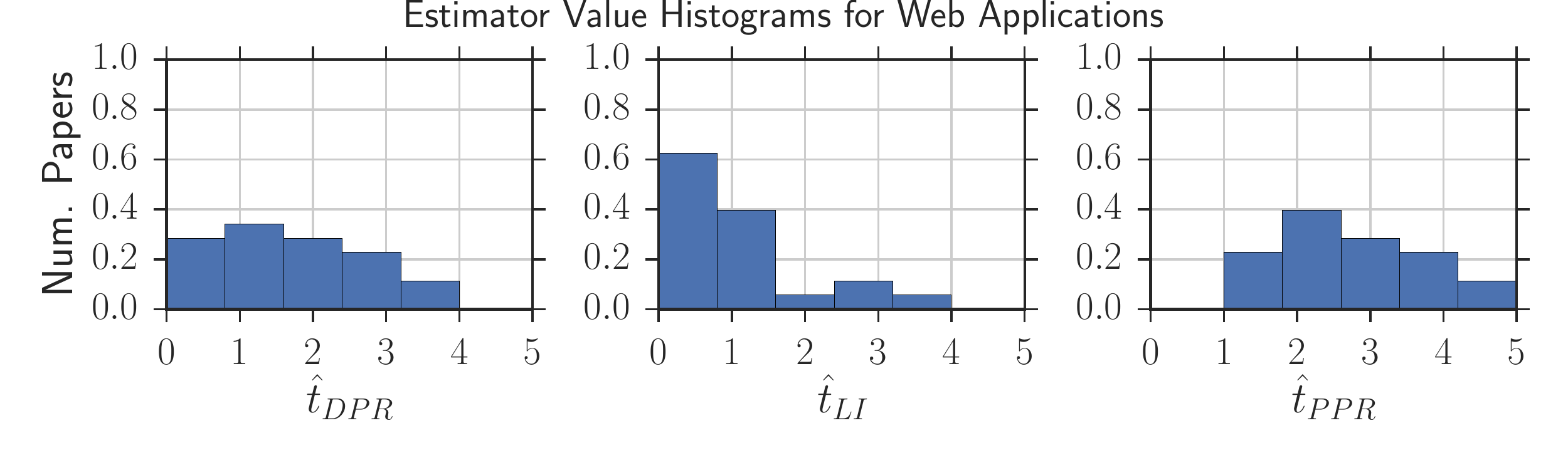}
\includegraphics[width=0.5\textwidth]{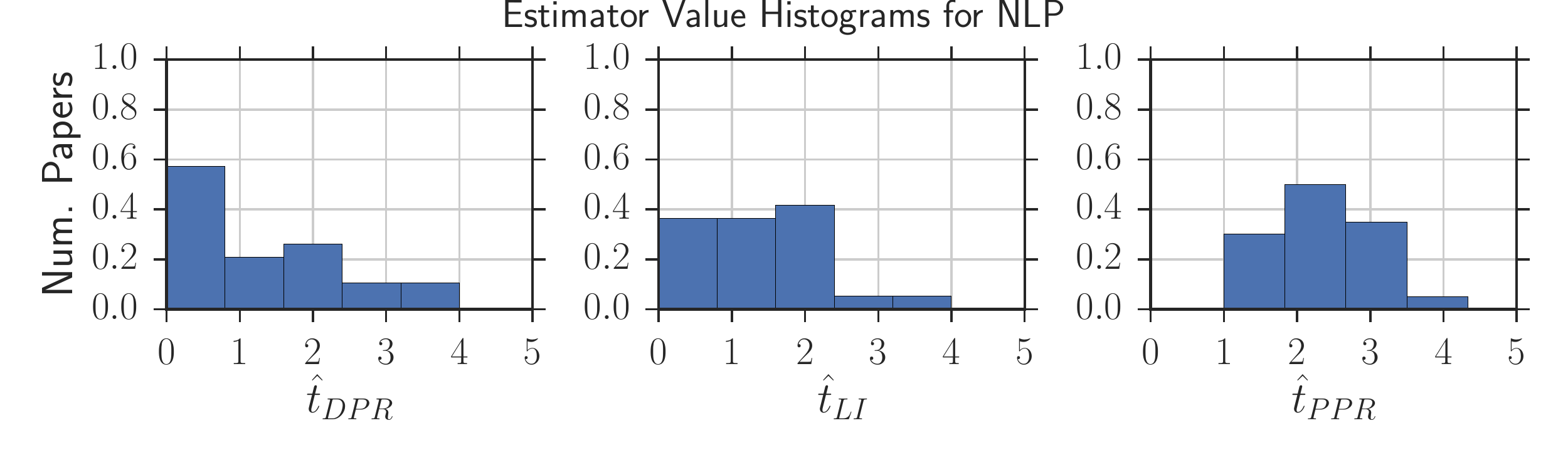}
\includegraphics[width=0.5\textwidth]{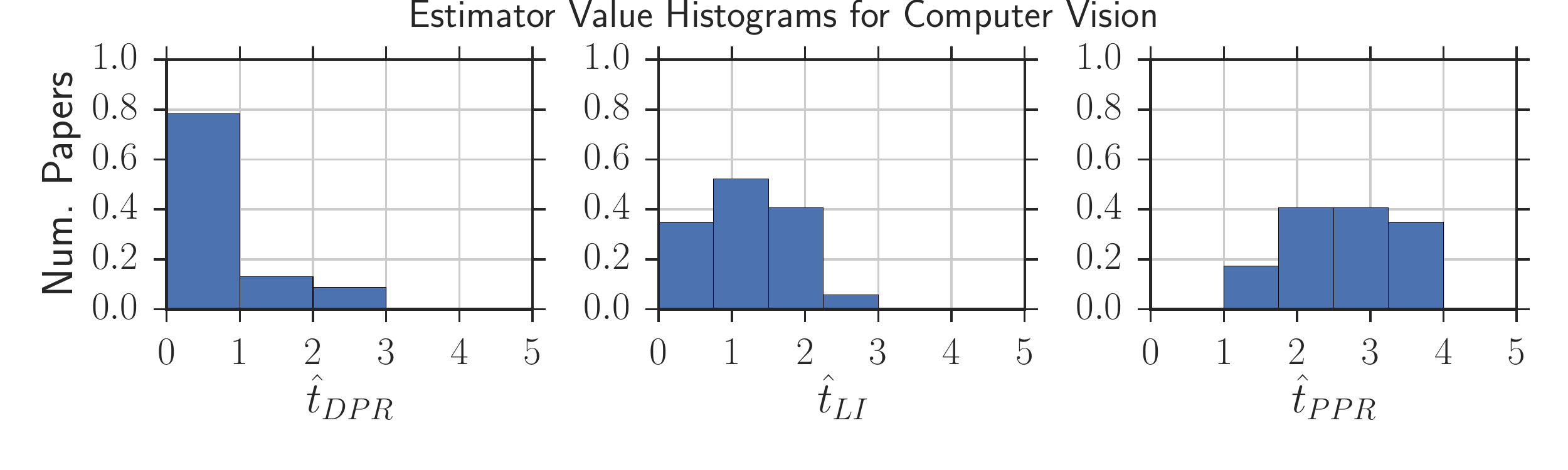}
\vspace{-18pt}
\caption{Distribution of number of iterations by workflow component.}
\label{fig:overallDistro}
\end{figure}

\begin{figure}[h]
\centering
\includegraphics[width=0.4\textwidth]{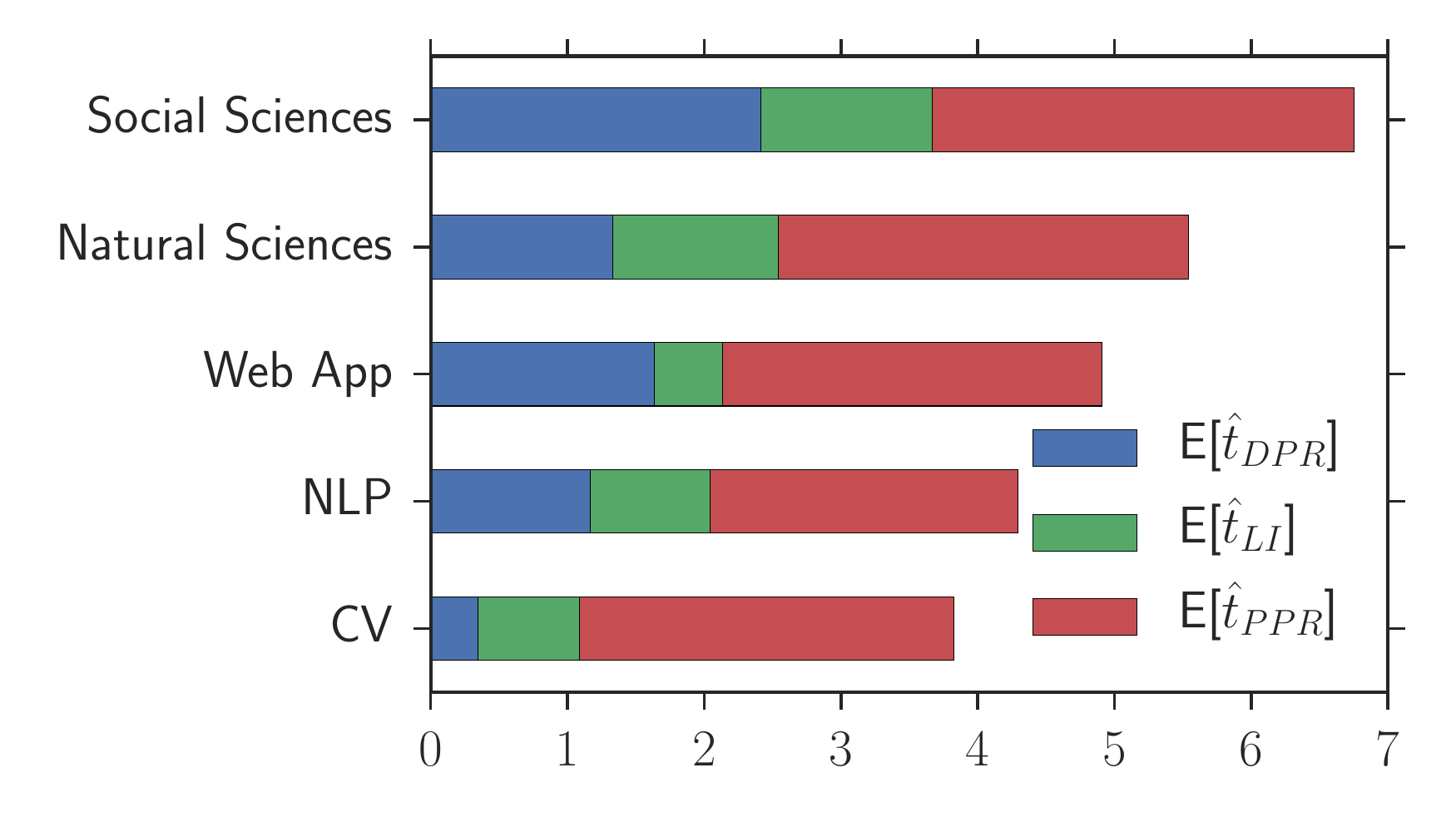}
\vspace{-20pt}
\caption{Mean iteration count by domains.}
\label{fig:meanCount}
\end{figure}

Figure~\ref{fig:overallDistro} shows 
the histograms for the three iteration estimators 
$\hat{t}_{DPR}, \hat{t}_{LI}, \hat{t}_{PPR}$
across the entire corpus (top row) and by domain (rows 2-6).
A bin in every histogram represents an integral value for the estimators,
and bin heights equal the fraction of papers with the bin value as their estimates.
The mean values for the estimators by domains are shown in the stacked bar chart in Figure~\ref{fig:meanCount}, 
where the total bar length is equal to the average number of iterations in each domain.
From these two figures, we see that
1) most papers use $\geq 1$ evaluation methods, 
evident from the fact that histograms in the third column in Figure~\ref{fig:overallDistro}
are skewed towards $\hat{t}_{PPR} \geq 2$;
2) PPR is the most common iteration type across all domains,
evident from the length of the E[$\hat{t}_{PPR}$] bars in Figure~\ref{fig:meanCount}; and
3) on average, more DPR iterations are reported than L/I iterations in every domain except computer vision,
as illustrated by the relative lengths of the E[$\hat{t}_{DPR}$] and E[$\hat{t}_{LI}$] bars in Figure~\ref{fig:meanCount}.

When grouped by domains, 
we see that the distributions for certain domains 
deviate a great deal from the overall trends in Figure~\ref{fig:overallDistro}.
Domains dominated by deep neural nets (DNNs),
which are designed to replace manual feature engineering for higher order features,
tend to skew towards fewer DPR and more L/I iterations,
such as NLP and CV.
Additionally, there are only a few highly processed datasets 
studied in all NLP and CV papers,
further reducing the need for data pre-processing in these domains.
On the other hand, social and natural sciences exhibit the opposite trend in the histograms in Figure~\ref{fig:overallDistro}, 
biasing towards more DPR iterations.
This is largely due to the fact that 
both domains rely heavily on domain knowledge to guide ML
and strongly prefer explainable models.
In addition, a large amount of data is required to enable training of DNNs.
The scale of data is often much smaller for SocS and NS than NLP and CV,
thus preventing effective application of DNNs and requiring more manual features.

\subsection{Data Pre-processing by Domain}

\begin{table*}[h]
\centering
\begin{tabular}{|c|c|c|c|c|}
\hline
\textbf{SocS} & \textbf{NS} & \textbf{WWW} & \textbf{NLP} & \textbf{CV} \\
\hline
\cellcolor{rgb5} Join   ($31.0\%$) & \cellcolor{rgb2}Feature def.  ($40.6\%$) & \cellcolor{rgb2}Feature def.  ($36.1\%$) & \cellcolor{rgb2}Feature def.  ($32.1\%$) & \cellcolor{rgb2}Feature def.  ($37.5\%$) \\
\hline
\cellcolor{rgb2}Feature def.   ($27.6\%$) &\cellcolor{rgb3} Univar. FS ($18.8\%$) & \cellcolor{rgb5}  Join ($22.2\%$)  & \cellcolor{c12}BOW ($17.9\%$) & \cellcolor{c12}BOW ($25.0\%$)\\
\hline
\cellcolor{rgb1}Normalize  ($17.2\%$) & \cellcolor{rgb1}Normalize  ($12.5\%$) & \cellcolor{rgb1} Normalize ($13.9\%$) & \cellcolor{rgb5}Join ($14.3\%$)  &\cellcolor{c4} Interaction  ($25.0\%$)\\
\hline
\cellcolor{rgb4}Impute  ($6.9\%$) & \cellcolor{c2}PCA  ($9.4\%$) & \cellcolor{c14}Discretize  ($8.3\%$) & \cellcolor{rgb1}Normalize  ($10.7\%$) & \cellcolor{rgb5}Join ($12.5\%$) \\
\hline
\end{tabular}
\caption{Common DPR operations ordered top to bottom by popularity. 
Join = joining multiple data sources;
Feat. def. = custom logic for fine-grained feature extraction;
Univer. FS = univariate feature selection, 
using criteria such as support and correlation per feature;
BOW = bag of words;
PCA = principal component analysis, 
a common dimensionality reduction technique.}
\label{tab:dpr}
\end{table*}

Table~\ref{tab:dpr} shows the most popular DPR operations in each application domain, 
ordered top to bottom by popularity,
with abbreviations expanded in the caption.
While the table reaffirms common knowledge such as feature normalization is important, 
Table~\ref{tab:dpr} also shows two striking results:
1) joining multiple data sources is common in four of the five domains surveyed;
2) $\frac{1}{3}$ of the papers contain fine-grained features defined using domain knowledge 
across all domains.
Result 1) suggest that unlike classroom and data competition settings 
in which the input data resides conveniently in a single file,
data in real-world ML applications is aggregated from multiple sources 
(e.g., user database and event logs).
Result 2) contradicts the common belief 
that ML applications have collectively progressed beyond handcrafted features 
thanks to the advent of deep learning (DL).
In addition to the incompatibilities with DL in some domains 
mentioned in Section~\ref{sec:iterCount},
the efficacy of features designed using domain knowledge 
versus using DL to search for the same features without domain knowledge 
is possibly another contributing factor.

\subsection{Learning/Inference by Domain}

\begin{table*}[h]
\centering
\begin{tabular}{|c|c|c|c|c|}
\hline
\textbf{SocS} & \textbf{NS} & \textbf{WWW} & \textbf{NLP} & \textbf{CV} \\
\hline
\cellcolor{rgb5}GLM ($36.0\%$) & \cellcolor{rgb1}SVM  ($32.7\%$)  & \cellcolor{rgb5}GLM  ($37.0\%$)  & \cellcolor{rgb3}RNN  ($32.4\%$) & \cellcolor{c12}CNN ($38.2\%$) \\
\hline
\cellcolor{rgb1}SVM   ($28.0\%$) & \cellcolor{rgb5}GLM  ($15.4\%$) &  \cellcolor{c14} RF   ($11.1\%$)   & \cellcolor{rgb5}GLM  ($14.7\%$) &\cellcolor{rgb1} SVM  ($17.6\%$) \\
\hline
\cellcolor{c14}RF     ($20.0\%$)  &\cellcolor{c14} RF    ($13.5\%$)   & \cellcolor{rgb1} SVM   ($11.1\%$) & \cellcolor{rgb1}SVM  ($11.8\%$) & \cellcolor{rgb3}RNN  ($17.6\%$) \\
\hline
\cellcolor{rgb2}Decision Tree  ($12.0\%$)    & \cellcolor{rgb4}DNN   ($13.5\%$)  & \cellcolor{c2}Matrix Factorization  ($11.1\%$) & \cellcolor{c12}CNN  ($8.8\%$) & \cellcolor{c14}RF  ($5.9\%$) \\
\hline
\end{tabular}
\caption{Common model classes ordered top to bottom by popularity per domain.
GLM = generalized linear models (e.g., logistic regression);
RF = random forest;
SVM = support vector machine;
R/CNN = recursive/convolutional neural networks.}
\label{tab:mc}
\end{table*}

Table~\ref{tab:mc} lists the most popular model classes for each application domain,
with abbreviations expanded in the caption.
We have already discussed the disparity between 
the popularity of DL in CV/NLP and other domains in Section~\ref{sec:iterCount}.
Most traditional approaches such as GLM, SVM, and Random Forest 
are still in favor with most domains,
since the large additional computation cost for DL 
often fails to justify the incremental model performance gain.
Matrix factorization, which is highly amenable to parallelization,
 is popular in web applications for supporting recommendation engines.
Interestingly, SVM is the most popular method in natural sciences by a large margin 
(100\% more popular than the second most popular option),
possibly due to its ability to support higher order functions through kernels. 
NS applications experimenting with DL are mostly computer vision related.

\begin{table*}[h]
\centering
\begin{tabular}{|c|c|c|c|c|}
\hline
\textbf{SocS}  & \textbf{NS}  & \textbf{WWW} & \textbf{NLP}  & \textbf{CV}\\
\hline
\cellcolor{rgb2}Regularize ($40.0\%$) & \cellcolor{rgb5}CV   ($31.8\%$) & \cellcolor{rgb2}Regularize  ($41.2\%$)  &\cellcolor{c14} LR  ($39.4\%$) & \cellcolor{c14}LR  ($46.2\%$) \\
\hline
\cellcolor{rgb5}CV    ($30.0\%$)         & \cellcolor{c14}LR  ($22.7\%$)     &  \cellcolor{c14} LR    ($23.5\%$)      & \cellcolor{rgb1} Batch size  ($24.2\%$) & \cellcolor{rgb1} Batch size  ($30.8\%$) \\
\hline
\cellcolor{c14}LR    ($10.0\%$)   & \cellcolor{rgb3}DNN arch.  ($18.2\%$)     &  \cellcolor{rgb1} Batch size  ($11.8\%$)  & \cellcolor{rgb3}DNN arch.  ($18.2\%$) & \cellcolor{rgb3}DNN arch.  ($11.5\%$) \\
\hline
\cellcolor{rgb1} Batch size  ($10.0\%$)    & \cellcolor{rgb4}Kernel  ($9.1\%$)   & \cellcolor{rgb5}CV  ($11.8\%$) & \cellcolor{rgb4}Kernel  ($6.1\%$) & \cellcolor{rgb2}Regularize  ($11.5\%$) \\
\hline
\end{tabular}
\caption{Most popular model tuning operations by domain.
CV = cross validation;
LR = learning rate;
DNN arch. = DNN architecture modification;
Kernel specifically applies to SVM.}
\label{tab:mt}
\end{table*}

Table~\ref{tab:mt} shows the most popular model tuning operations by domains.
The top two operations, learning rate and batch size, 
are both concerned with the training convergence rate,
suggesting that training time is an important factor in all domains.
Cross validation and regularization are both mechanisms 
to control model complexity and overfitting to observed data.
Lower complexity models usually result in faster inference time
and better ability to generalize to more unseen data.

\subsection{Post Processing by Domain}

\begin{table*}[h]
\centering
\begin{tabular}{|c|c|c|c|c|}
\hline
\textbf{SocS} & \textbf{NS} & \textbf{WWW} & \textbf{NLP} & \textbf{CV} \\
\hline
\cellcolor{rgb5}P/R  ($25.7\%$)              &  \cellcolor{rgb2} Acc.  ($28.6\%$)  &\cellcolor{rgb2} Acc.  ($20.8\%$)   & \cellcolor{rgb5}P/R  ($29.2\%$) & \cellcolor{c14}Vis.  ($33.3\%$) \\
\hline
\cellcolor{rgb2}Acc.  ($20.0\%$)           & \cellcolor{rgb5}P/R    ($18.6\%$)    &   \cellcolor{rgb5}P/R  ($20.8\%$)       &\cellcolor{rgb2} Acc.  ($27.1\%$)  &\cellcolor{rgb2}Acc.  ($29.8\%$)  \\
\hline
\cellcolor{c11}Feat. Contrib.  ($17.1\%$)     &\cellcolor{c14} Vis.   ($15.7\%$)  &\cellcolor{rgb3}  Case ($13.2\%$)  & \cellcolor{rgb3}Case  ($14.6\%$) & \cellcolor{rgb5}P/R  ($17.5\%$) \\
\hline
\cellcolor{c14}Vis.  ($14.3\%$)    & \cellcolor{rgb4}Correlation  ($11.4\%$)          & \cellcolor{c12}DCG   ($9.4\%$)           &\cellcolor{c2} Human Eval.  ($8.3\%$) &\cellcolor{rgb3} Case  ($12.3\%$) \\
\hline
\end{tabular}
\caption{Most popular evaluation methods by domain.
P/R = precision/recall;
Acc. = accuracy;
Vis. = visualization;
Feat. Contrib. = feature contribution to model performance;
NCG = discounted cumulative gain, popular in ranking tasks;
Case = case studies of individual results.}
\label{tab:metrics}
\end{table*}

Of the evaluation methods listed in Table~\ref{tab:metrics},
P/R, accuracy, correlation, and DCG are summary evaluations of model performance
while case study, feature contribution, human evaluation, and visualization 
are fine-grained methods towards insights to improve upon the current model.
While the former group can be used automatically such as in grid search,
the latter group is aimed purely for human understanding.

\subsection{System Desiderata}
\label{sec:props}
%!TEX root=main.tex

The results in Section~\ref{sec:results} suggest a number of properties that a versatile and effective human-in-the-loop ML system should possess:
\begin{denselist}
\item \textbf{Iteration.} 
Developers iterate on their workflows in every application domain 
and test out changes to all components of the workflow. 
Understanding the most frequent changes 
helps us develop systems 
that anticipate and respond rapidly to iterative changes.
\item \textbf{Fine-grained feature engineering.} 
Handcrafted features designed using domain knowledge is still an indispensable part of the workflow development systems in all domains 
and should therefore be adequately supported instead of dismissed as an outdated practice.
\item \textbf{Efficient joins.} 
Data is often pooled from multiple sources, thus requiring systems to support efficient joins in the data pre-processing component.
\item \textbf{Explainable models.}
Many domains have yet to embrace deep learning due to their needs for explainable models.
The system should provide ample support to help developer interpret model behaviors.
\item \textbf{Fast model training.}
The fact that the most tuned model parameters are related to training time suggests 
that developers are in need of systems that have fast model training, but also low latency for the end-to-end workflow execution in general.
\item \textbf{Fine-grained results analysis.}
Fine-grained and summary evaluation methods are equally popular across all domains.
Thus, model management systems should provide support for not only summary metrics 
but also more detailed model characteristics.
\end{denselist}

\noindent We are in the process of developing a system,
titled \name~\cite{dorx2017}, that is aimed at accelerating
iterations in human-in-the-loop ML workflow development,
using many of the properties listed above as guiding principles.

\section{Conclusion and Future Work}
\label{sec:conclusion}
%!TEX root=main.tex

We conduct a statistical study on
 the iterative development process 
for ML applications in multiple domains.
Our approach involves collecting carefully designed statistics 
from applied machine learning literature 
in order to reconstruct the iterative 
process that led to the results reported.
We present our survey findings across domains
and discuss desired ML system properties 
as suggested by the trends discovered from our survey data.
The statistics and estimators described in our work 
can be further developed into a benchmark
for systems specifically designed 
to address human-in-the-loop ML needs.

\bibliographystyle{ACM-Reference-Format}
\bibliography{sigproc}

\balance

\end{document}